# An Automated Compatibility Prediction Engine using DISC Theory Based Classification and Neural Networks

Chandrasekaran Anirudh Bhardwaj
VIT Chennai
canirudh.bhardwaj2014@vit.ac.in

Megha Mishra
VIT Chennai
megha.mishra2014@vit.ac.in

Sweetlin Hemalatha
VIT Chennai
sweetlin.hemalatha@vit.ac.in

**ABSTRACT**

Traditionally psychometric tests were used for profiling incoming workers. These methods use DISC profiling method to classify people into distinct personality types, which are further used to predict if a person may be a possible fit to the organizational culture. This concept is taken further by introducing a novel technique to predict if a particular pair of an incoming worker and the manager being assigned are compatible at a psychological scale. This is done using multilayer perceptron neural network which can be adaptively trained to showcase the true nature of the compatibility index. The proposed prototype model is used to quantify the relevant attributes, use them to train the prediction engine, and to define the data pipeline required for it.

**Keywords**
DISC, Psychology, Datamining, Neural Networks, Compatibility Prediction

## INTRODUCTION

Traditionally large organizations have leveraged the uses of psychometric test to classify people into different personality types to determine if they fit the organizational culture. Recently, small scale organizations and new startups also have started to realize the need of psychometric based classification. DISC [1] theory is a model proposed by William Moulton Marston, and is used extensively in the industry for such classification of personalities.

DISC theory is widely used in the industry [2] due to its easy administration as a test, which can be of the form of a formal test or an informal test. Formal testing conditions include a questionnaire format and basic case studies. Informal testing includes approaches such as gamification using websites [3] or using datamining techniques [4].

DISC theory stands for the four main types of personalities in which the individuals can be placed in, chiefly the dominant personality type, influence personality type, steady personality type and compliance personality type [5]. The DISC based classification first divides the clients into two containers called active and passive types. The active types are typically confrontational, and are more ambitious than other types. They are not easily satiated, and try to climb up the corporate ladder [6]. The passive type is more likely to not stand up for themselves as often as the active types, and are generally satisfied with their positions, and do not possess the desire to go out of the way to achieve their goals. The containers are further divided into sub containers called task and people oriented categories. Figure 1 explains how the personality types are split. The people associated with the task sub type are more inclined to be more ready to accept challenging tasks, but are more likely to be unfriendly, and correspondingly the people associated with the people subtype are assumed to be more people friendly and less ready to take tasks which may challenge their current position.

|  | TASK | PEOPLE |
|---|---|---|
| ACTIVE | DOMINANT | INFLUENCE |
| PASSIVE | COMPLIANT | STEADY |

**Figure 1. Categorization of personality types**

The people who fit the active type and task subtype are classified as belonging to the dominant category, while people who fit the active type and people subtype are termed as belonging to the influence category. People who fit the passive type and belong to the task subtype are classified as compliant personality, while people who belong to the people subtype are classified as belonging to the steady category. In excess to these classifications, organizations typically take into account the scores derived from a standardized test such as Activity Vector Analysis [7], Eysenck Personality Questionnaire [8], and Big Five Personality Traits [9], which represents the different parameters which try to represent the person. The scores are then used as filtration criterion to eliminate candidates. This method is used because the tests are usually bias free, and there is nearly no possibility of discrimination in terms of race or gender, due to their standardized nature.

**LITREATURE REVIEW**

Automated systems using simple mathematical formulae have been proposed [10,11, 12] in the past, but the main pitfall with such systems is the lack of elasticity of the models. The hypothesis proposed for such approaches are quite rigid, whereas a neural network based model is malleable and is capable of adapting to circumstances locally as well as globally.

Personality based classification of potential employees and as team members has been used for predicting if the potential candidate would fit in a group, and would they be productive or not. The associated business value of a candidate is often measured by the psychometric traits possessed by an individual [13]. The leadership of an organization is expected to be possessing some quality traits [14] such as and active type in their personality, as to proactively guide the organization to success [15].

DISC based systems have also been used in other applications such as for customized teaching for students [11], and for customizing treatment methods of dental patients [16] etc.

Teams that are balanced in terms of personality types have been found to be statistically more productive than teams that do not [17]. The proposed prototype model uses synthetic data created with a hypothesis that a level (n+1) employee would like to balance their team by adding people of level (n) in the corporate ladder, who possess diametrically opposite traits as them. Here, the levels indicate the corporate ladder, and higher levels indicate upper echelons of management. This hypothesis is modeled as a synthetic dataset, which is used to pre-train a neural network model, which will help predict the probability that the level (n) candidate would be a fit to the team or not.

A majority of the corporations use psychometric classification tasks to eliminate unsuitable candidates, and do not leverage the full potential of psychometric assessments. The proposed method takes this method one step further, and tries to optimize the psychometric compatibility at an individual scale by using machine learning concepts such as neural networks. These algorithms can be fit using existing data and then used to predict the compatibility of future recruits and their immediate manager. The objective of the predictions is to aid the process of forming optimal groups [18] for maximal productivity.

**PROPOSED MODEL**

For training the prediction engine, test scores obtained from a standardized test [6] are used. The scores are represented using six attributes namely faith, decisiveness, adaptability, dominance, ambition and emotional management. The structure of this feature vector is shown in figure 2.

The test scores are scaled to range between 0 and 10 units. This is due to the fact that the neural networks are scale variant. The score of both the recruit and the manager being assigned are taken into account and concatenated together to form an input vector of dimension twelve features. The output is binary vector which gives one if both the people are compatible or else zero if the compatibility is less.

**Figure 2. Structure of input features for an individual**

| Faith | Decisiveness | Adaptability | Dominance | Ambition | Emotional Management |
|---|---|---|---|---|---|
|  |  |  |  |  |  |

The synthetic data is formed by creating a dataset, in which the data points are filled according to discreet uniform distribution. This is to ensure that the data is consistent with the real life values. The distribution is centered to a value of five (The middle value) in each attribute. The values situated inside each feature vary between zero and ten. The optimum point is assumed to be the complement of the vector associated with the individual [17]. The optimum can be represented as

$$Y(i,j) = 10 - X(i,j)$$

Where, i represents the identifier index for the individual and j represents the attribute number for each individual.

The Euclidean distance from the optimum point for each individual to every other individual is calculated. This metric signifies the deviance of the point versus the optimum. If the point is far from the optimum, it means the point is incompatible and vice versa. If the point is situated very close to optimum, then it means the individual has a high probability of compatibility with their manager. This distance is represented as $D(Y, X)$. The Maximum of all the minimum distances for each individual is calculated for the dataset, to serve as the cutoff point. This is done to ensure that each individual at least has minimum of one compatible choice. This minimum cutoff is represented as $CutoffDist$.

The points which are less than the $CutoffDist$ are initially assumed to satisfy the compatibility criterion. This can be represented as

If ( $D(Y,X) \leq CutoffDist$),

    then assign $Compatiblity\ (Y, X) = 1$

else

    assign $Compatibility(Y, X) = 0$

This is used to obtain a dataset which could be used to pre-train the model, which could then further be used to train using real world data. This dataset is termed binarized utility dataset. Now, the data is extracted from the matrix, choosing one pair of individuals at a time and comparing their compatibility. This data generated from the matrix can be ingested as input data for the prediction engine.

Input data is of the shape containing 12 attributes, representing the test scores of the two people for whom the compatibility is to be tested.

**EXPERIMENT AND RESULT DISCUSSION**

The output is a binary variable which represents if the pair of people are compatible or not. The model could also be trained to predict the probability of compatibility. The prediction engine model is a multi-layer perceptron network [19] consisting of 4 layers with 64 nodes each. The model is trained until the accuracy reaches convergence. The structure is represented in figure 3.

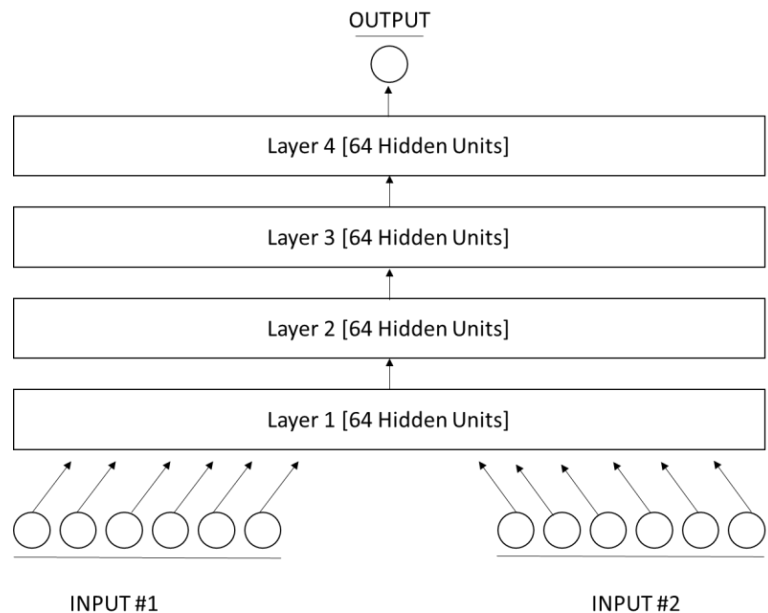

**Figure 3. Structure of the multilayer perceptron neural network**

The nodes are connected densely with each other, meaning each node in layer (n-1) is connected to each node in layer (n). The data was split into training and testing data in a ratio of 8:2 with 20% of the data as test data and the remaining 80% as training data. The training data was further divided into training data and validation data in a ratio of 8:2. As result of double splitting of the training data, the actual data used to train the neural network is only 64% of the actual dataset size.

On fitting the model to the training data until convergence of weights, the accuracy of the test data tested on the prediction engine was found to be 99.84%. The precision, recall and f1-scores obtained for the test data is reflected in table 1.

**Table 1. Classification Report for multilayer perceptron neural network**

| Class | Precision | Recall | F1-Score | Support |
|---|---|---|---|---|
| 1 | 1.00 | 1.00 | 1.00 | 98.901% |
| 0 | 0.90 | 0.87 | 0.89 | 1.098% |

Here, the class label 0 represents the value of the binary output when the pair of two individuals are not compatible with each other, and a value of 1 in the class label represents that the pair of two individuals are compatible on a psychological level.

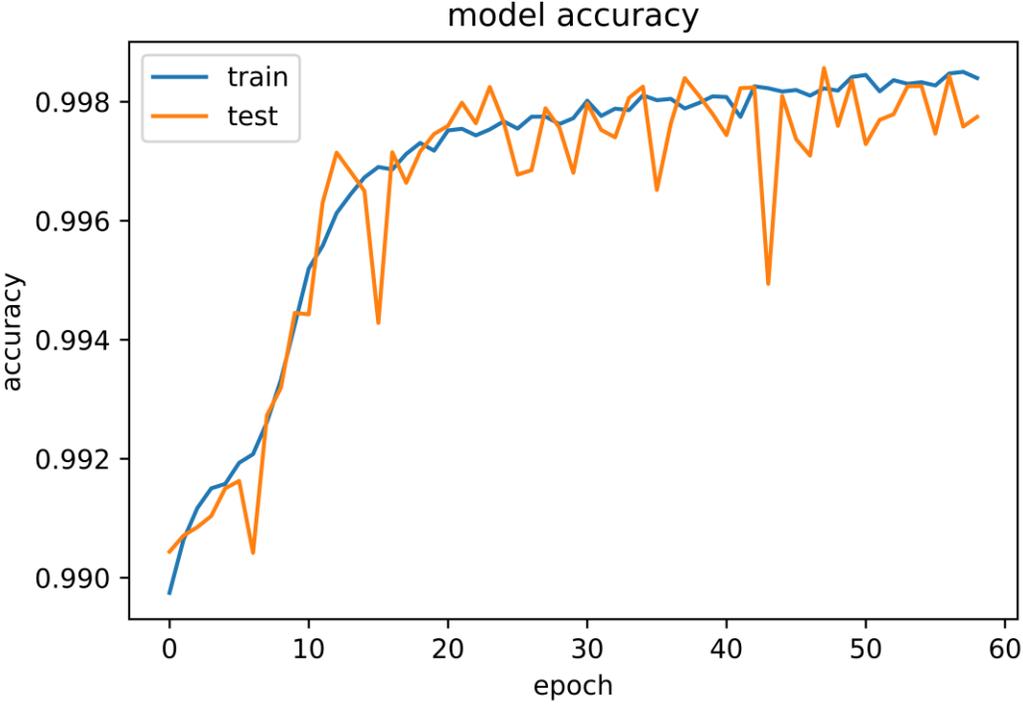

**Figure 4. shows the convergence of model accuracy over training and validation data**

The code was run for 59 epochs, after which the validation loss converged to a stabilized value. This can be visualized in Figure 4, which shows the convergence achieved by the prediction engine model's accuracy metric. The loss function used for updating the weights of the neural network was binary cross-entropy. The convergence of the loss function can be seen in Figure 5.

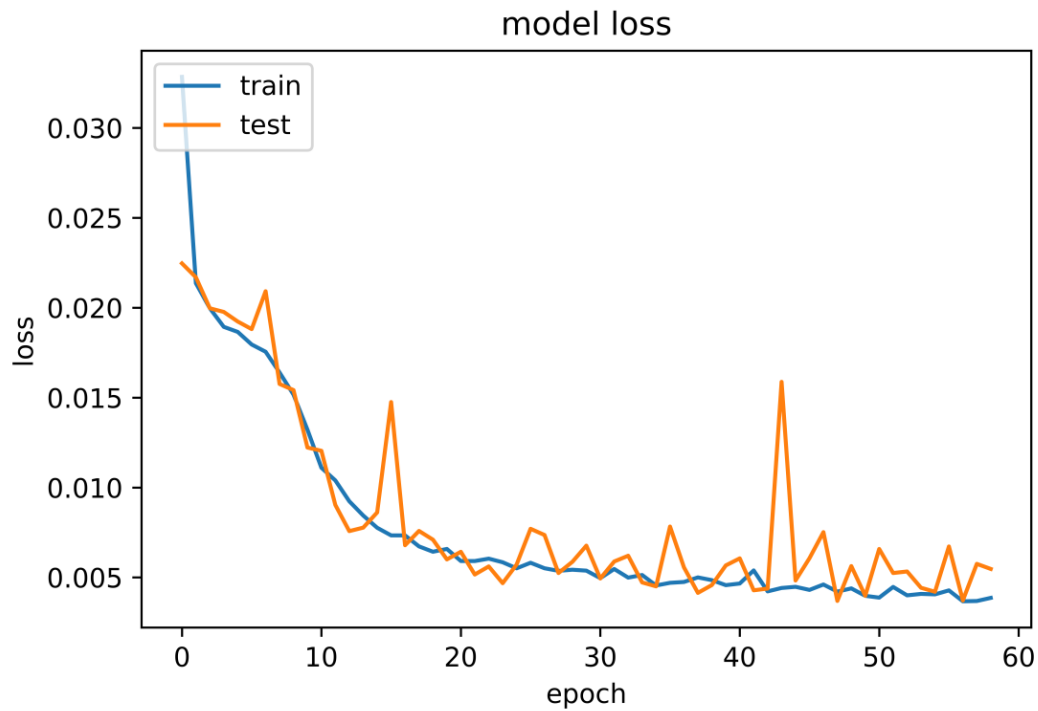

**Figure 5. shows the convergence of loss function over training and validation data**

**CONCLUSION**

The neural network performs exceedingly well in identifying and learning linear/non-linear patterns, and relationships. This behavior is reflected in the precision, recall and f1 scores for the dataset shown in table 1. The multilayer perceptron neural network is a simple model which performs the needed work in a fast manner as compared to other neural network frameworks like recurrent neural networks or convolution neural networks.

The learning capability of a multilayer perceptron neural network model is represented in the accuracy and precision score exhibited by the model. Even though the problem is of the form of a class imbalance problem, the multilayer perceptron neural networks works very well, and has a good precision, recall and f1-score.

Furthermore, the model can be fit to real world data to mimic complex patterns which will enable the neural network to perform well in predicting the compatibility of the individuals.

## FUTURE WORK

Various advances in neural networks have been made, and the complexity of the networks could be decreased by adding convolution nodes instead of simple perceptron nodes, as they reduce the amount of updates required to the node weights to achieve accuracy.

Other approaches such as including unsupervised learning with self-organizing maps to understand how the machine learns to classify the individuals based on their skillsets into different personality categories could be beneficial to reduce misclassification error.

Extreme learning can be used in future research to further reduce the training time for the algorithm and to reduce the number of updates to the nodes. The structure of the nodes for the artificial neural networks, as well as more features could be added to the input vectors to increase the efficiency.